%% file: main.tex
\documentclass{article}
\usepackage{spconf,amsmath,graphicx}
\usepackage{amsfonts}

\usepackage{booktabs}
\usepackage{tabularx}
\usepackage{xcolor}
\usepackage{enumitem}

\newcommand{\unpublished}{\cite{shillingfordLargeScaleVisualSpeech2019, serdyukAudioVisualSpeechRecognition2021,
changConformerAllYou2024,sonLipReadingProfile2017, makinoRecurrentNeuralNetwork2019}}

\def\presec{\vspace{-5pt}}
\def\postsec{\vspace{-5pt}}
\def\presub{\vspace{-10pt}}
\def\postsub{\vspace{-5pt}}

\title{Enhancing CTC-Based Visual Speech Recognition}
\name{Hendrik Laux$^{1}$, Anke Schmeink$^{1}$}

\address{
$^1$Chair of Information Theory and Data Analytics, RWTH Aachen University, Germany
}

\begin{document}
\ninept

\maketitle
%
\begin{abstract}
\presec
This paper presents LiteVSR2, an enhanced version of our previously introduced efficient approach to Visual Speech Recognition (VSR). Building upon our knowledge distillation framework from a pre-trained Automatic Speech Recognition (ASR) model, we introduce two key improvements: a stabilized video preprocessing technique and feature normalization in the distillation process. These improvements yield substantial performance gains on the LRS2 and LRS3 benchmarks, positioning LiteVSR2 as the current best CTC-based VSR model without increasing the volume of training data or computational resources utilized.
Furthermore, we explore the scalability of our approach by examining performance metrics across varying model complexities and training data volumes. LiteVSR2 maintains the efficiency of its predecessor while significantly enhancing accuracy, thereby demonstrating the potential for resource-efficient advancements in VSR technology.

\end{abstract}


\section{Introduction}
\postsec
\label{sec:intro}

Visual Speech Recognition (VSR), also known as automatic lip reading, is the process of decoding linguistic content from sequences of visual input depicting a speaker's facial movements, particularly those of the lips, teeth, and tongue. While sharing fundamental objectives with Automatic Speech Recognition (ASR), VSR distinguishes itself by operating solely on the visual input domain, as opposed to the acoustic signals utilized in ASR.
VSR models try to learn a mapping between visemes and a sequence of characters or subword tokens, where visemes represent the visual counterparts of phonemes in spoken language. However, the exclusive use of visemes in speech transcription introduces inherent ambiguities that pose a significant challenge in VSR systems. For instance, the bilabial consonants /p/, /b/, and /m/ may appear visually indistinguishable, necessitating sophisticated and complex models able to infer additional information from the linguistic context \cite{mayerLabialVisemeReconsidered2011}.
Generally, VSR systems employ a pipeline that includes face detection, facial landmark localization, region of interest extraction (typically the mouth area), feature extraction, and finally, sequence modeling for text generation. Modern approaches leverage deep learning architectures such as 3D Convolutional Neural Networks (CNNs) for spatio-temporal feature extraction, followed by sequence models like Recurrent Neural Networks (RNNs) \cite{shillingfordLargeScaleVisualSpeech2019} or Transformers \cite{maLiRALearningVisual2021} for temporal modeling.
VSR can be extended to Audio-Visual Speech Recognition (AVSR), where both visual and acoustic modalities are jointly processed. AVSR systems aim to leverage the complementary nature of these inputs, potentially improving robustness in noisy acoustic environments or in scenarios with visual occlusions \cite{serdyukAudioVisualSpeechRecognition2021}. 

Despite the conceptual similarities between ASR and VSR, the latter presents substantially higher computational demands during model training and inference. While ASR typically processes two-dimensional Mel spectrograms, VSR must handle high-dimensional spatio-temporal data in the form of video frame sequences. This increase in input dimensionality not only elevates memory requirements but also significantly extends training and inference times.
The disparity in available training data further exacerbates the challenges in VSR development. While ASR benefits from extensive, publicly accessible speech corpora such as LibriSpeech (1000 hours) [1] and Common Voice (over 20,000 hours across multiple languages) [2], the data landscape for VSR is considerably more constrained. Many of the datasets employed in recent state-of-the-art VSR models remain proprietary or undisclosed, often due to complex copyright issues associated with video data. Public research is largely confined to the "Lip Reading Sentences in the Wild" (LRS) series of datasets \cite{chungLipReadingSentences2017,afourasLRS3TEDLargescaleDataset2018}, which, while indispensable for VSR research, are limited in scale compared to their audio counterparts.

\presub
\subsection{Related Work}
\postsub

Soon after the publication of the first end-to-end lip reading model LipNet \cite{assaelLipNetEndtoEndSentencelevel2016}, the field has achieved significant progress towards narrowing the gap between ASR and VSR performance, with state-of-the-art models quickly beating professional human lip-readers \cite{shillingfordLargeScaleVisualSpeech2019, chungLipReadingSentences2017}. Since then, results on the common LRS2 and LRS3 benchmarks have constantly and rapidly improved over the years. Among the most significant causes for improved results are modern sequence processing architectures such as Transformers \cite{shiLearningAudioVisualSpeech2022,vaswaniAttentionAllYou2017} and Conformers \cite{gulatiConformerConvolutionaugmentedTransformer2020, maEndToEndAudioVisualSpeech2021}, the use of auto-regressive sequence decoders \cite{maLiRALearningVisual2021, prajwalSubwordLevelLip2022, maVisualSpeechRecognition2022} or sophisticated visual backbones such as Vision Transformers \cite{prajwalSubwordLevelLip2022, changConformerAllYou2024}. 
The inherent similarity between ASR and VSR suggests the potential for leveraging ASR models in the training of VSR models, an approach that has been increasingly adopted in recent literature \cite{maLiRALearningVisual2021, yeoAKVSRAudioKnowledge2024a, ahnSyncVSRDataEfficientVisual2024}.
Furthermore, the increasing power and availability of modern deep learning hardware and the abundance of raw data on video streaming platforms have allowed the research field to scale up the resource-intensive task of training VSR models, achieving even better results. However, this comes at the cost of significant resource requirements and the use of non-public audio-visual datasets \unpublished. 
The trend of upscaling has opened the door for research on efficient VSR, attempting to train models with less data and computational resources  \cite{ahnSyncVSRDataEfficientVisual2024}, however until now, these models still fall significantly short of the state-of-the-art. 
In our preceding publication \cite{lauxLiteVSREfficientVisual2024}, we introduced the LiteVSR framework, which addressed these resource constraints through an innovative approach to VSR model training. LiteVSR employs a novel knowledge distillation technique leveraging pre-trained ASR models and re-using parts of this model at the same time, facilitating the training of a resource-efficient architecture achieving competitive performance using only publicly available datasets. 

\presub
\subsection{Contribution}
\postsub

In this paper, we present LiteVSR2, an evolution of our previous work on efficient visual speech recognition. The contributions of this work are as follows:
\begin{itemize}[leftmargin=*] \vspace{-3pt}
    \item We introduce feature normalization techniques to stabilize training and improve the efficacy of knowledge transfer.\vspace{-3pt}
    \item We present a stabilized video preprocessing pipeline allowing for more effective training of the visual base.\vspace{-3pt}
    \item We empirically show that our pre-training objective aligns well with the goal of reducing error metrics on standard benchmarks.\vspace{-3pt}
    \item We provide an analysis of our approach's scalability, examining the impact of increased computational resources and varying amounts of labeled training data on performance.\vspace{-3pt}
\end{itemize}

\noindent Our updated model demonstrates significant performance improvements over our previous work. We achieve substantial performance gains for VSR models trained solely on unlabeled data, pushing the boundaries of unsupervised learning in this domain. Furthermore, we demonstrate marked improvements when training with limited labeled data, achieving the best metrics for a CTC-based model on both LRS2 and LRS3 benchmarks with as little as 59 hours of labeled training data. 

\presec
\section{Methodology}
\label{sec:methodology}
\postsec

We use the LiteVSR framework from \cite{lauxLiteVSREfficientVisual2024} to distill knowledge from a robust ASR model into a model for Visual Speech Recognition. For better comparability, we keep most of the model and data hyperparameters fixed to those of the original LiteVSR model. Thus, the following sections will briefly summarize the model architecture, data preprocessing and training details including the changes made for the experimental evaluation in this work. 

\presub
\subsection{Model Architecture}
\postsub

We divide the pre-trained ASR model into two components: a lower section, the audio base $B_a$, and an upper section, the audio head $H_a$. For our experimental work, we employ the CTC-based \textit{stt\_en\_conformer\_ctc\_small} model from the Nvidia NeMo toolkit~\cite{kuchaievNeMoToolkitBuilding2019}. This model transcribes Mel spectrogram audio into a byte-pair-encoded (BPE) alphabet of size 1025 \cite{sennrichNeuralMachineTranslation2016}, using two convolutional subsampling layers and a 17-layer Conformer stack. For our experiments, we split the model after the 8th Conformer layer. Consequently, $B_a$ comprises the convolutional subsampling layers and around half of the Conformer stack, while $H_a$ consists of the remaining half of the Conformer layers and the linear decoder. 
During the pre-training stage, we use the audio base to infer features of shape $[N, T, d\,]$ ($N$: batch size, $T$: maximum sequence length in the batch, $d$: model dimension of the Conformer stack) from the audio signal $x_a$. As these features capture high-level information about the input speech, we train our visual base $B_v$ to align with these features while using the corresponding silent frames $x_v$ as an input. This alignment is achieved through a distance-based encoding loss denoted as: $\mathcal{L}_{\text{enc}} = dist(B_{a}(\mathbf{x}_{a}), B_{v}(\mathbf{x}_{v}))$. 

\begin{figure}[h]
\centering 
\includegraphics[width=0.85\linewidth]{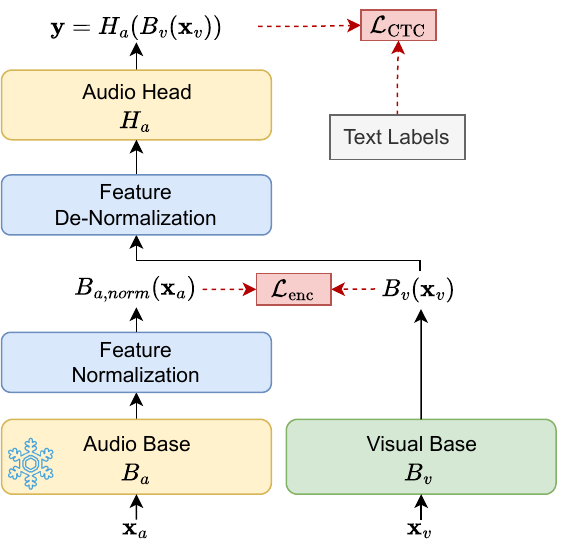}
\vspace{-10pt}
\caption{Updated LiteVSR Architecture with Feature Normalization.}
\label{fig:architecture} 
\vspace{-10pt}                              
\end{figure}

The adaptation of the visual feature space towards the audio feature space enables us to employ the (hitherto unmodified) audio head $H_a$ to transcribe the output of the visual base into text. This approach facilitates end-to-end lip reading from silent frames without relying on labeled data during training. Subsequently, using labeled data, the model $H_a(B_v(x_v))$ can be fine-tuned to further enhance prediction accuracy. 
For the experiments conducted in this work, we use the visual base architecture from \cite{lauxLiteVSREfficientVisual2024}, comprising two 3D-convolutional layers and a ResNet18 for the visual backbone, followed by a stack of 12 Conformer layers with $d=256$.  We choose the Mean Square Error (MSE) as the encoding loss and the Connectionist Temporal Classification loss (CTC) \cite{gravesConnectionistTemporalClassification2006} for the subsequent fine-tuning. For a more comprehensive discussion on the general LiteVSR approach, we direct readers to \cite{lauxLiteVSREfficientVisual2024}.

\presub
\subsection{Feature Normalization}
\postsub

\begin{figure}[h]
\centering 
\includegraphics[width=1\linewidth]{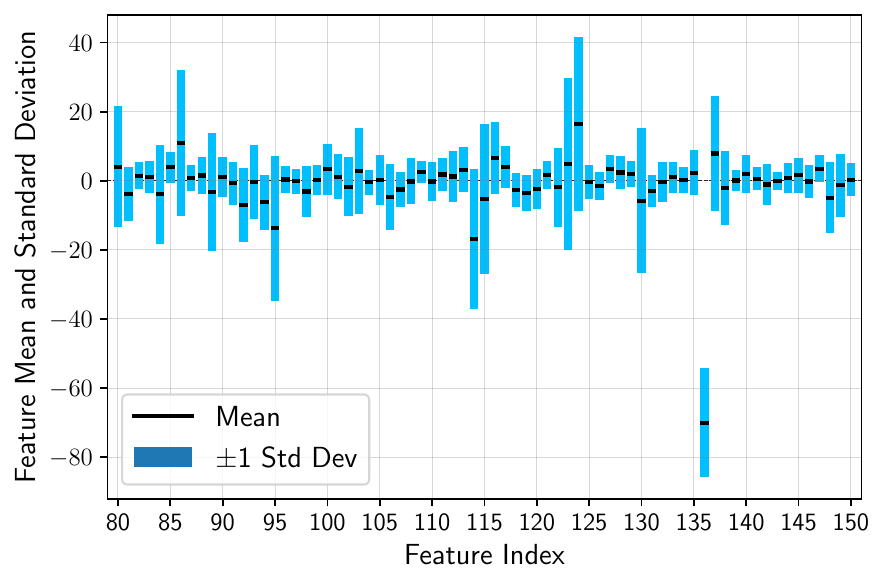}
\vspace{-20pt}
\caption{Audio feature statistics for the LRS2-pretrain dataset. The figure shows the mean (black lines) and standard deviation (blue bars) of features produced by the 8th Conformer layer of the \textit{stt\_en\_conformer\_ctc\_small} model. These outputs from $B_a$ are used as targets for training $B_v$. For readability, only a subset of the 176 features is shown}
\label{fig:featstats} 
\end{figure}

\begin{figure}[t]
\centering 
\includegraphics[width=1\linewidth]{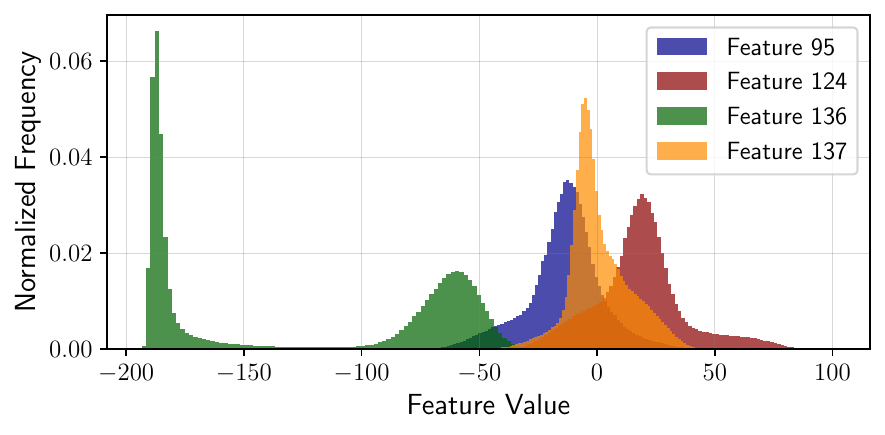}
\vspace{-20pt}
\caption{Distribution of selected features from Figure~\ref{fig:featstats}.}
\label{fig:feathist} 
\vspace{-10pt}                              
\end{figure}

The baseline model from \cite{lauxLiteVSREfficientVisual2024} demonstrates competitive results with short training times. However, we observe exceptionally large gradients and occasional instabilities during the pre-training stage. A detailed examination of the encodings $B_a(x_a)$ produced by the audio base reveals a highly disproportionate scaling of individual features within the feature space. 
Figure~\ref{fig:featstats} shows the mean and standard deviation of different features after the 8th Conformer layer of the \textit{stt\_en\_conformer\_ctc\_small} model. We can observe that the scale, variance and offset of features are highly dissimilar. While many of them are centered around a mean of zero with single-digit standard deviations, there are noteworthy exceptions across the whole feature space, with the most prominent being feature number 136. This anomaly is not specific to this model and layer, as we can observe similar patterns in other models (e.g. the \textit{large} and \textit{medium} variant of the \textit{stt\_en\_conformer\_ctc} model) and split layers in the same model. Figure~\ref{fig:feathist} provides further insight into the unique characteristics of feature 136. It is the only feature among all 176 that exhibits a bimodal distribution, while all other features (more or less) share similarities with a Gaussian distribution. 

The aforementioned characteristics of the target feature space introduce several challenges in training the visual base. Employing a distance-based loss function results in exceptionally large loss values and gradients, potentially leading to numerical instabilities, particularly when utilizing half or mixed precision training. While the adoption of advanced precision formats such as \textit{bfloat16} \cite{burgessBfloat16ProcessingNeural2019} mitigates this issue to some extent, these formats are exclusively available on the most recent GPUs and TPUs and maintaining gradients within reasonable bounds remains a prudent approach regardless of the training environment.
Furthermore, the heterogeneous scaling across features introduces a problem of disproportionate feature weighting. Features exhibiting large variances contribute disproportionately high individual loss values when employing distance-based loss functions such as the Mean Squared Error (MSE). This imbalance can potentially skew the learning process, causing the model to overly prioritize certain features.

To address these issues, we propose normalizing each individual feature based on its observed statistical properties. This approach serves to mitigate the problem of large-scale gradients and the resulting instabilities and imbalances. By standardizing the feature space, we aim to create a more uniform learning landscape, potentially improving the stability and effectiveness of the training process. 

\setlength{\abovedisplayskip}{4pt}
\setlength{\belowdisplayskip}{4pt}

\setlength{\belowcaptionskip}{4pt}

Let $\boldsymbol{\mu} \in \mathbb{R}^d$ and $\boldsymbol{\sigma} \in \mathbb{R}^d$ represent the mean and standard deviation vectors respectively for each of the $d$ individual features produced by $B_a$. We define the normalization operation $\mathcal{N}$ for the features $\mathbf{f} \in \mathbb{R}^{N \times T \times d}$ as: 
$$\mathcal{N}(\mathbf{f}) = (\mathbf{f} \ominus \boldsymbol{\mu}) \oslash \boldsymbol{\sigma},$$
where the subtraction and division operations are applied element-wise across the feature dimension.
Correspondingly, we define the de-normalization operation $\mathcal{N}^{-1}$ as:
$$\mathcal{N}^{-1}(\mathbf{f}_\text{norm}) = \mathbf{f}_\text{norm} \odot \boldsymbol{\sigma} \oplus \boldsymbol{\mu},$$
with element-wise sum and multiplication.
Normalization is applied to the output of $B_a$ during training and de-normalization is used on the output of $B_v$ during inference and labeled fine-tuning, ensuring consistency in the feature space. Specifically, the encoding loss becomes:
$$\mathcal{L}_\text{enc} = \text{dist}(\mathcal{N}(B_a(\mathbf{x}_a)), B_v(\mathbf{x}_v)).$$
With $B_v$ now inferring normalized features, the output of $B_v$ has to be de-normalized before being passed to $H_a$, as the audio head expects the original feature properties of the pre-trained ASR model:
$$\hat{\mathbf{y}} = H_a(\mathcal{N}^{-1}(B_v(\mathbf{x}_v))).$$
This normalization approach standardizes the feature space, mitigating the issues of disproportionate scaling and large gradients observed in the original formulation. The updated architecture in Figure~\ref{fig:architecture} illustrates the integration of Feature (De-)Normalization into the LiteVSR framework.

\presub
\subsection{Input Video Processing}
\postsub

In our previous work, we utilized dlib \cite{kingDlibmlMachineLearning2009} to detect 68 facial landmarks in every input video frame. The mouth center is estimated by averaging the (x,y)-coordinates of four key landmarks, while the mouth width is calculated as the Euclidean distance between the left and right corners of the lips. However, we can observe that the dynamic changes in mouth shape during speech (e.g., the protrusion when pronouncing an "o" sound) leads to rapid variations in the mouth's width and, consequently, in the crop region. These fast dynamics significantly impact the feature extraction capabilities of the 3D-convolutional backbone employed in our model.

To address this issue, we evaluated two alternative cropping methods: fixed cropping and smooth cropping. Both methods begin by obtaining the mouth's (x,y)-center position and Euclidean width for each individual frame. The fixed cropping approach selects the maximum width and average center position across the entire video, applying these parameters consistently to all frames to ensure a stable crop region and size. In contrast, the smooth cropping method applies Gaussian smoothing ($\sigma=4$) to the centers and widths, thereby reducing, but not entirely removing, the inter-frame variability of crops throughout the video.
Our empirical evaluation revealed that both smooth and fixed cropping methods yield significant, but equal performance improvements over the frame-wise cropping technique. Given the similar outcomes, we opt for the fixed cropping method due to its computational simplicity and ease of implementation.
To enhance the robustness of our model, we maintain our previous augmentation techniques during training. These include randomly displacing the sample's center coordinate and introducing random variations to the sample's mouth width, thereby increasing the diversity of the training data.

\presub
\subsection{Data}
\postsub

We restrict the data used to train our baseline models to publicly available datasets, specifically the LRS2 datasets \cite{chungLipReadingSentences2017}, featuring material from several BBC formats, as well as the LRS3 dataset \cite{afourasLRS3TEDLargescaleDataset2018}, created from TED and TEDx presentation videos. In the pre-training stage, we use the unlabeled pre-train sets provided in both datasets with a total amount of 639 hours of video material. The fine-tuning stage uses 59 hrs of labeled audio-visual data from the labeled parts of the LRS2 (main) and LRS3 (trainval) datasets. 
For our scale-up experiments we additionally extract a subset of the unlabeled VoxCeleb2 dataset \cite{chungVoxCeleb2DeepSpeaker2018}, from which we choose all videos in English language with a maximum duration of 6 seconds, yielding a total of 350hrs of audio-visual data. We create pseudo-labels for the videos using the OpenAI Whisper \textit{medium.en} ASR model \cite{radfordRobustSpeechRecognition2022}.

\presub
\subsection{Training Details}
\postsub

For all training runs, we employ an Adam Optimizer \cite{adam2014} with hyperparameters $\beta_{1} = 0.9$, $\beta_{2} = 0.98$, and $\epsilon = 10^{-9}$, using a mini-batch size of 64. During the pre-training phase, we implement a Noam learning rate schedule \cite{vaswaniAttentionAllYou2017} with 50,000 warmup steps and a peak learning rate of $8 \times 10^{-4}$. The relatively high learning rate is facilitated by the Feature Normalization technique, which enhances training stability.
For the fine-tuning phase using labeled data, we adjust the learning rate schedule to warm up over 10,000 steps, with a reduced peak rate of $1 \times 10^{-4}$ to accommodate the additional CTC loss. Both pre-training and fine-tuning processes are conducted on a single NVIDIA A100 GPU with bfloat16 precision \cite{burgessBfloat16ProcessingNeural2019}.

\presec
\section{Experimental Evaluation}
\label{sec:results}
\postsec

In this section, we present the results achieved using the proposed updates to our previous approach. We place our results in the context of other, relevant VSR publications using a CTC-based decoding scheme. Further, we provide an analysis on how well our training objective of minimizing the encoding loss aligns with the downstream objectives of minimizing the CTC loss and, ultimately, the Word Error Rate.    

\presub
\subsection{Benchmark Results}
\postsub

Table~\ref{tab:comparison} presents a comparison of CTC-based methods evaluated on the common LRS benchmarks. Our approach, which incorporates Feature Normalization in the knowledge distillation process and employs the enhanced frame preprocessing scheme, demonstrates substantial improvements over our previous model iteration. Using exclusively unlabeled audio-visual data, we achieve absolute reductions in Word Error Rate (WER) of 5.1\% and 7.1\% on the LRS2 and LRS3 test sets respectively (relative reductions of 10.8\% and 13\%). When using limited amounts of labeled data for fine-tuning the pre-trained model, we can lower the WER on LRS2 and LRS3 by 3.4\% and 7.6\% (relative reduction of 9.7\% and 16.6\%). Comparing with the related works from Table~\ref{tab:comparison}, our fine-tuned model achieves the best Word Error Rates among all CTC-based Visual Speech Recognition models on both the LRS2 and LRS3 benchmarks using just 59 hrs of publicly available labeled data for training. 

The incorporation of 350 hours of pseudo-labeled English data from the VoxCeleb2 dataset results in a further reduction of the WER to 36.7\% on the LRS3 dataset. However, this additional data did not yield improvements for the LRS2 dataset. Increasing the image resolution to 96x96 pixels during unlabeled pre-training yields WERs of 40.6\% and 47.4\% for LRS2 and LRS3, respectively. While this change requires approximately twice the video memory, it demonstrates the potential for performance enhancement through increased computational resources and data volume.

\input{results_table}

\presub
\subsection{Alignment of the Pre-Training Objective}
\postsub

A key consideration in our approach is the relationship between the pre-training objective of minimizing the encoding loss and the downstream metrics of CTC loss and Word Error Rate. In the context of unlabeled training, direct optimization of the CTC loss is not feasible due to the absence of transcriptions. Similarly, WER, being a non-differentiable metric, cannot be directly optimized at all. We thus analyze whether our chosen pre-training objective effectively serves as a proxy for these downstream metrics.
Figure~\ref{fig:encscatter} provides empirical evidence of this alignment. The scatter plots and trend lines demonstrate a strong positive correlation between the encoding loss and both the CTC loss and WER on the LRS2 and LRS3 test sets. This suggests a successful indirect optimization of the CTC loss and WER through our pre-training objective and validates our approach of training VSR models with unlabeled data only. 
\input{encloss_figures}

\presec
\section{Conclusion}
\label{sec:conclusion}
\postsec

LiteVSR2 maintains its predecessor's focus on computational efficiency while achieving state-of-the-art performance among CTC-based VSR models on standard benchmarks, demonstrating the potential for resource-conscious advancements in VSR technology. We further show the scalability of our approach, yielding improved performance when leveraging increased computational resources and larger datasets. 
Our analysis reveals a strong alignment between our pre-training objective and relevant downstream VSR metrics. This correlation validates our approach to distill knowledge from ASR models, indicating a better prediction of encodings to directly translate into enhanced VSR performance.

\clearpage

\bibliographystyle{IEEEbib}
{\footnotesize
\bibliography{VSR_clean}
}

\end{document}

%% file: results_table.tex
\begin{table}[t]
\centering
\resizebox{\columnwidth}{!}{%
\begin{tabular}{@{}l rr rr@{}}
& \multicolumn{2}{c}{\textbf{Data Quantity [hrs]}} & \multicolumn{2}{c}{\textbf{WER [\%]}} \\
\cmidrule(lr){2-3} \cmidrule(lr){4-5}
\textbf{Method} & \textbf{unlabeled} & \textbf{labeled} & \textbf{LRS2} & \textbf{LRS3} \\
\midrule
Shillingford et al. \cite{shillingfordLargeScaleVisualSpeech2019} \ & - & 3,886 & - & 55.1 \\
Afouras et al. \cite{afourasASRAllYou2020} & 334 & 699 & 51.3 & - \\
Afouras et al. \cite{afourasASRAllYou2020} & 334 & 475 & - & 59.8 \\
LiRA \cite{maLiRALearningVisual2021} & 439 & 225 & 38.8 & - \\
LiRA \cite{maLiRALearningVisual2021} (reported by \cite{shiLearningAudioVisualSpeech2022}) & 433 & 30 & - & 72.8 \\
LiRA \cite{maLiRALearningVisual2021} (reported by \cite{shiLearningAudioVisualSpeech2022}) & 1,759 & 433 & - & 58.4 \\
AV-HuBERT \cite{shiLearningAudioVisualSpeech2022} & 1,759 & 30 & - & 40.7 \\
AV-HuBERT \cite{shiLearningAudioVisualSpeech2022} & 1,759 & 433 & - & 38.6 \\
\midrule
LiteVSR \cite{lauxLiteVSREfficientVisual2024} & 639 & 59 & 35.0 & 45.7 \\
\midrule
\textbf{LiteVSR2} & 639 & 59 & \textbf{31.6} & 38.1 \\
\textbf{LiteVSR2} & 639 & 409 & 32.0 & \textbf{36.7} \\
\midrule
\midrule
LiteVSR \cite{lauxLiteVSREfficientVisual2024} & 639 & 0 & 47.4 & 54.7 \\
\midrule
\textbf{LiteVSR2} & 639 & 0 & 42.3 & 47.6 \\
\textbf{LiteVSR2-96x96} & 639 & 0 & \textbf{40.6} & \textbf{47.4} \\
\bottomrule
\end{tabular}%
}
\caption{Comparison of state-of-the-art VSR models using pure CTC decoding, trained on labeled data (upper part) and unlabeled data only (lower part)}
\label{tab:comparison}
\vspace{-10pt}

\end{table}

%% file: encloss_figures.tex
\begin{figure}[t]

\begin{minipage}[b]{.99\linewidth}
  \centering
  \centerline{\includegraphics[width=\linewidth]{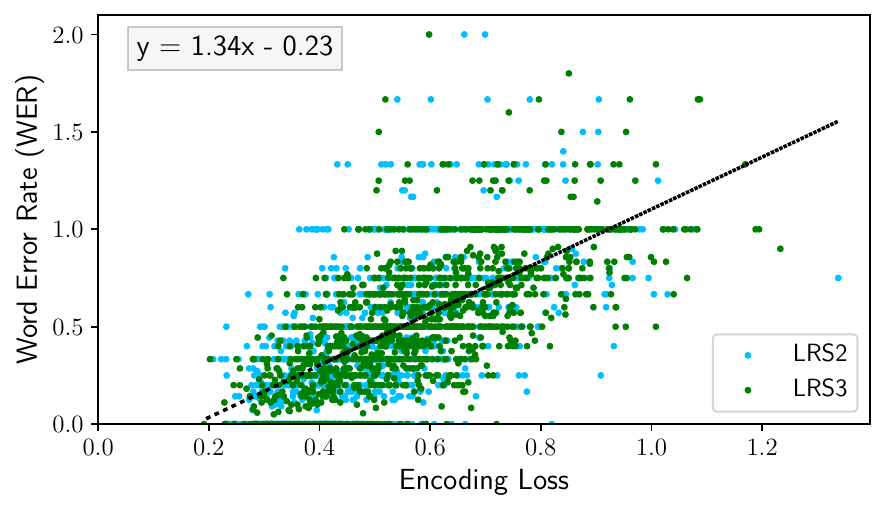}}
\end{minipage}
\hfill
\begin{minipage}[b]{0.99\linewidth}
  \centering
  \centerline{\includegraphics[width=\linewidth]{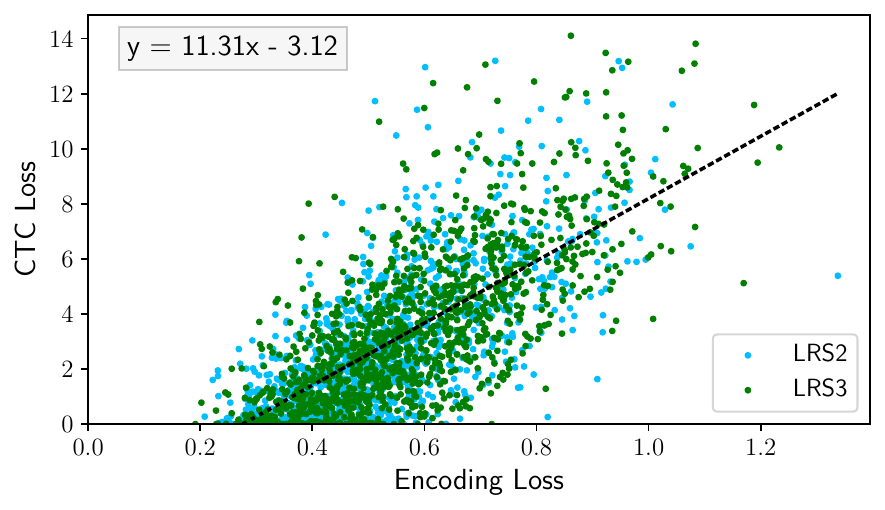}}
\end{minipage}
\vspace{-10pt}
\caption{Scatter plot showing the relation between the encoding loss $\mathcal{L}_{\text{enc}}$ and the WER metric (upper plot) / the CTC loss (lower plot) after pre-training. We feed the silent frames of each video to the pre-trained visual base to obtain the visual feature representation $B_v(\mathbf{x}_v)$. We use these features to obtain the encoding loss between audio and visual features and let $H_a$ transcribe the visual features to calculate the CTC loss and WER for each sample of the (unseen) LRS2 and LRS3 test sets. The black line indicates the trend-line obtained from using linear regression on the data.}
\label{fig:encscatter}
\end{figure}